\documentclass[a4paper,10pt]{llncs}
\pdfoutput=1
\usepackage{hyperref}
\usepackage{amssymb}
\usepackage{amstext}
\usepackage{amsmath}
\usepackage[pdftex]{graphicx}

\title{\uppercase{Quantum-like Uncertain Conditionals for Text Analysis}}
\author{Alvaro Francisco Huertas-Rosero \inst{1} and C. J. van Rijsbergen\inst{2}\\
  \{alvaro, keith\}@dcs.gla.ac.uk}
\institute{University of Glasgow \and University of Cambridge}
\toctitle{Uncertain Conditional}

\begin{document}

\maketitle

\begin{abstract}
 Simple representations of documents based on the occurrences of terms are ubiquitous in areas like Information Retrieval, and also frequent in Natural Language Processing.  In this work we propose a logical-probabilistic approach to the analysis of natural language text based in the concept of Uncertain Conditional, on top of a formulation of lexical measurements inspired in the theoretical concept of ideal quantum measurements.  The proposed concept can be used for generating topic-specific representations of text, aiming to match in a simple way the perception of a user with a pre-established idea of what the usage of terms in the text should be.  A simple example is developed with two versions of a text in two languages, showing how regularities in the use of terms are detected and easily represented.
 \end{abstract}

\section{Introduction}\label{sec:introduction}

How do prior expectations/knowledge affect the way a user approaches a text, and how they drive the user's attention from one place of it to another? This is a very important but tremendously complex question; it is indeed as complex as human perception of text can be.  Including such effects in the representation of text may be a relatively easy way to enhance the power of a text retrieval or processing system.   In this work we will not address the question, but assume a simple answer to it, and follow it while building theoretical concepts that can constitute natural language text for retrieval of similar processing tasks.

The key concept to be defined will be an \textbf{Uncertain conditional} defined between lexical measurements, which will allow us to exploit structures and features from both Boolean and Quantum logics to include certain features in a text representation.

Automatic procedures for acquiring information about term usage in natural language text can be viewed as lexical measurements, and can be put as statements such as [term $t$ appears in the text]\footnote{In this paper we will use the convention that anything between square brackets [ and ] is a proposition}, to which it is possible to assign true/false values.  These can be regarded as a set of \textbf{propositions}. Some relations between propositions have the properties of an \textbf{order relation} $\sqsubseteq$: for example, when one is a particular case of the other, e.g $P_1 = $ [term ``research'' appears in this text] and $P_2 = $ [term ``research'' appears in this text twice] we can say that $P_2 \sqsubseteq P_1$ or that $P_2$ is \textit{below} $P_1$ according to this ordering.

The set of propositions ordered by relation $\sqsubseteq$ can be called a \textit{lattice} when two conditions are fulfilled \cite{Lattices}: 1) a proposition exists that is above all the others (\textbf{supremum}), and 2) a proposition exists that is below all the others (\textbf{infimum}).  When any pair of elements of a set has an order relation, the set is said to be \textbf{totally ordered}, as is the case with sets of integer, rational or real numbers and the usual order ``larger or equal/smaller or equal than '' $\geqslant/\leqslant$.  If there are pairs that are not ordered, the set is \textbf{partially ordered}.

Two operations can be defined in a lattice: the \textbf{join} [$A \land B$] is the higher element that is below $A$ and $B$  and the \textbf{meet} [$A \lor B$] is the lower element that is above $A$ and $B$.  In this work, only lattices where both the join and the meet exist and are unique.  These operations are sometimes also called conjunction and disjunction, but we will avoid these denominations, which are associated with more subtle considerations elsewhere \cite{QuantumConjDisj}.

In terms of only ordering, another concept can be defined: the \textbf{complement}.  Whe referring to propositions, this can also be called \textbf{negation}.  For a given proposition $P$, the complement is a proposition $\lnot P$ such that  their join is the supremum $sup$ and their meet is the infimum $inf$:
 \begin{equation}\label{eq:ComplementOrdering}
 	[P \land \lnot P = inf ] \land [ P \lor \lnot P = sup]
\end{equation}

Correspondences between two ordered sets where orderings are not altered are called \textbf{valuations}.  A very useful valuation is that assigning ``false'' or ``true'' to any lattice of propositions, where \{``false'',``true''\} is made an ordered set by stating [``false'' $\sqsubseteq$ ``true''].  With the example it can be checked that any sensible assignation of truth to a set of propositions ordered with $\sqsubseteq$ will preserve the order.  Formally, a valuation $V$ can be defined:
\begin{equation}
	V:\{P_i\} \rightarrow \{Q_i\}, \text{ such that } (P_i \sqsubseteq_P P_j) \Rightarrow (V(P_i) \sqsubseteq_Q V(P_j))
\end{equation}
where $\sqsubseteq_P$ is an order relation defined in $\{P_i\}$ and $\sqsubseteq_Q$ is an order relation defined in $\{Q_i\}$.  Symbol $\Rightarrow$ represents material implication: $[X \Rightarrow Y]$ is true unless $X$ is true and $Y$ is false.

Another very important and useful kind of valuations is that of \textbf{probability measures}: they assign a real number between 0 and 1 to every proposition.

 Valuations allow for a different way of defining the negation or complement: for a proposition $P$, the complement $\lnot P$ is such that in any valuation $V$, when $P$ is mapped to one extreme of the lattice (supremum $sup$ or infimum $inf$) then $\lnot P$ will be mapped to the other
 \begin{equation}\label{eq:ComplementValuation}
 	[[V(P) = sup] \iff [V(\lnot P) = inf]]\land [[V(\lnot P) = sup] \iff [V(P) = inf]]
\end{equation}
For Boolean algebras, this definition will be equivalent to that based on order only (\ref{eq:ComplementOrdering}), but this is not the case for quantum sets of propositions.

A lattice and a valuation can be completed with a way to assess if a process to use some propositions to infer others is correct.  The rules that have to be fulfilled by these processes are called \text{rules of inference}.  In this work we do not aim to assessing the correctness of a formula, but define instead a probability measure for relations [$A \hspace{3pt}\boxed{R}\hspace{3pt} B$].  So we will not be exactly defining some kind of logic but using something that formally resembles it.  The kind of logic this would resemble is \textit{Quantum Logic}, which will be explained next.

\subsection{Conditionals in Quantum Logics}

The description of propositions about objects behaving according to Quantum Mechanics have posed a challenge for Boolean logics, and it was suggested that the logic itself should be modified to adequately deal with these propositions \cite{QuantumLogics}.  Von Neumann's proposal was to move from standard propositional systems that are isomorphic to the lattice of subsets of a set (distributive lattice \cite{Lattices}), to systems that are instead isomorphic to the lattice of subspaces of a Hilbert subspace (orthomodular lattice \cite{BeltramettiQL}).

\begin{figure}[h!]
	\begin{center}
		\includegraphics[width=8cm]{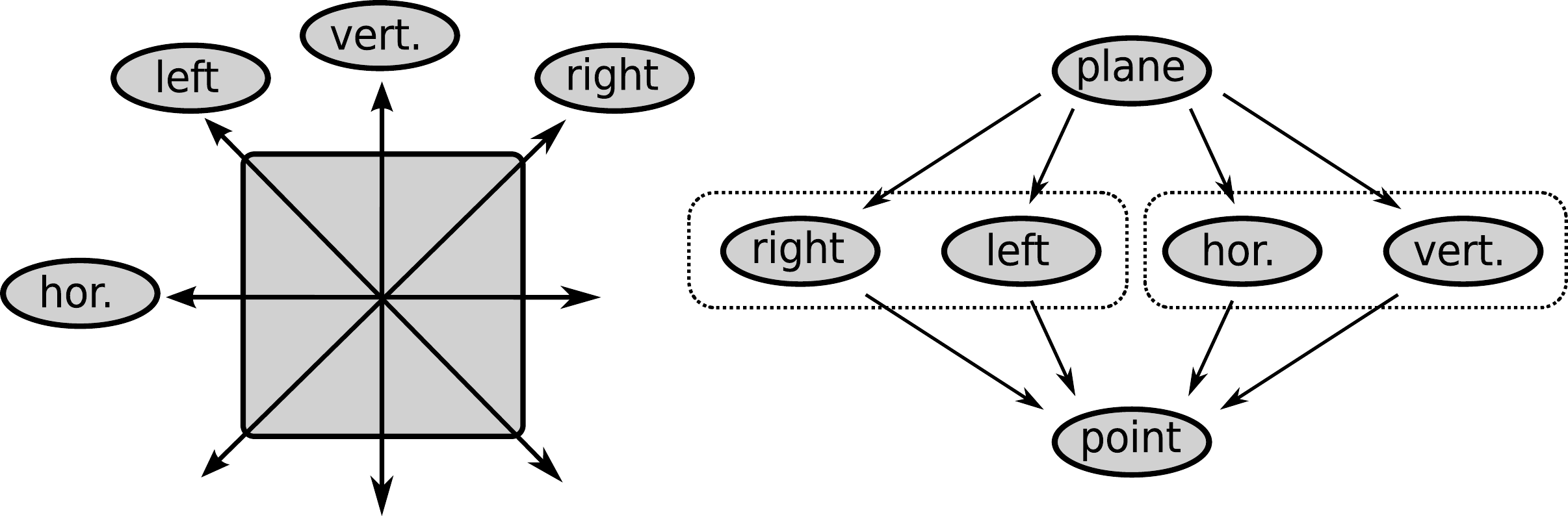}
	\end{center}
	\caption{System of propositions regarding the polarisation of a photon.  On the left, spaces representing the propositions.  On the right, order relations between them, represented with arrows. Subsets of orthogonal (mutually excluding) propositions are shown enclosed in dotted boxes.}
	\label{fig:Polarisation}
\end{figure}

A concept that is at the core of de difference between Boolean and Quantum structures is that of compatibility.   Quantum propositions may be incompatible to others, which means that, by principle, they cannot be simultaneously verified.  A photon, for example, can have various polarisation states, which can be measured either as linear polarisation (horizontal and vertical) or circular (left or right) but not both at a time: they are \textbf{incompatible} measurements.  The propositions about a particular polarisation measure can be represented in a 2D space as two pairs of perpendicular lines $\{\{[H],[V]\},\{[L],[R]\}\}$, as is shown in figure \ref{fig:Polarisation}.  The lattice of propositions would be completed with the whole plane $[plane]$ and the point where the lines intersect $[point]$.  The order relation $\sqsubseteq$ is ``to be contained in'', so $[point]$ is contained in every line, and all the lines are contained in the $[plane]$.

The fact that the measurements are pairwise exclusive is not reflected in the lattice itself, but in the kind of valuations that are allowed: when $[H]$ is true, $[V]$ can only be false, but neither $[L]$ nor $[R]$ are determined.  This can be described with valuation into a 3-element totally ordered set $\{false \sqsubseteq non-determined \sqsubseteq true \}$, together with two rules: 1) [only one proposition can be mapped to ``true'' and only one to ``false''] and 2) [if one proposition from an orthogonal pair is mapped to ``non-determined'', the other has to be mapped to ``non-determined'' as well].

The rudimentary formulation of valuation rules given in the example can be, of course, improved, which can be done using a geometrical probability measure.  According to Gleason's theorem \cite{GleasonsTheorem} this probability measure can be defined by choosing a set of orthogonal directions in space with weights that sum up to 1 $\{ w_i,\vec{e}_i\}$, with weights that sum up to one, and computing the weighted average of the fraction of these vectors that lies within each considered subset\footnote{This is not the standard formulation of the Quantum probability measure, but is entirely equivalent}, as follows:
\begin{equation}\label{eq:ProbMeasureProjector}
	V(\Pi) = \sum w_i \frac{||\Pi \vec{e}_i ||}{||\vec{e}_i||}
\end{equation}
The weighted orthogonal set $\{ w_i,\vec{e}_i\}$ is entirely equivalent to what is called \textbf{density operator} $\rho$ and equation (\ref{eq:ProbMeasureProjector}) is equivalent to the trace formula $V_\rho(\Pi) = Tr(\Pi \rho)$.

The valuations suggested in the example can be obtained by taking two of the orthogonal polarisations as $\vec{e}_1$ and $\vec{e}_2$ and interpreting probability 1 as ``true'', probability 0 as ``false'' and intermediates as ``non-determined''.

Defining conditionals in an orthomodular lattice has been a much discussed issue \cite{ReallyQLgardner71,ReallyQLpavicic04}, and this paper does not aim to contribute to the polemic; however, we will consider two aspects of the problem from the perspective of practical applicability: the role of valuation in the definition of a logic, and the role of complement or negation.

\subsubsection{Conditionals and the Ramsey Test}
Material implication $A \rightarrow B = \lnot A \lor B$ is known to be problematic when requirements other than soundness are considered (like relevancy \cite{RelevantLogic}, context\cite{SituationsHuibersBruza}, etc.) and other kinds of implication are preferred in areas like Information Retrieval \cite{Rijsbergen1984}.   A key issue in the consideration of an implication is what is the interpretation of $[A \rightarrow B]$ when $A$ is false. One possible approach to this issue is to consider ``what if \textit{it were} true'', which amounts to adopting \textit{counterfactual} conditional.  If we are interested in a probability measure rather than a true/false valuation, we may as well evaluate how much imagination do we need to put into the ``what if'' statement: how far it is from the actual state of things.  This is an informal description what is called \textbf{the Ramsey test} \cite{RamseyTestGardenfors}.  A simplified version of the Ramsey test can be stated as follows:
\begin{quote}
	To assess how acceptable a conditional $A \rightarrow B$ is given a state of belief, we find the least we could augment this state of belief to make antecedent $A$ true, and then assess how acceptable the consequent $B$ is given this augmented state of belief.
\end{quote}
In this work we will interpret \textit{state of beliefs} as a restriction of the set of possible valuations (including probability measures) that we will use to characterise a system of propositions: in the case of a purely Quantum formulation, it would mean imposing condition on the weighted orthogonal sets.  We will adopt a similar interpretation for lexical measurements in the next section.

\subsection{Uncertain Conditional and Text Analysis}

It has been suggested that high-level properties of natural language text such as topicality and relevance for a task can be described by means of conditional (implication) relations \cite[chapter 5]{Rijsbergen2004}, giving rise to a whole branch of the area of Information Retrieval devoted to logic models \cite{Lalmas1998}, \cite[chapter 8]{Widdows2004}.  In this work we will focus on the detection of patterns in the use of words that can also be put as implication-like relations.

In this work we will focus on lexical measurements as propositions, and will adopt the concept of \textbf{Selective Eraser} (SE) as a model for lexical measurements \cite{Erasers}.  A SE $E(t,w)$ is a transformation on text documents that preserves text surrounding the occurrences of term $t$ within a distance of $w$ terms, and erases the identity of tokens not falling within this distance.

A norm $|\cdot|$ for documents $D$ is also defined, that counts the number of defined tokens (can be interpreted as remaining information).  Order relations, as well as Boolean operations, can be defined for these transformations, and the resulting lattices are known to resemble those of Quantum propositions.

Order relations between SEs are defined for a set of documents $\{D_i\}$ as:
\begin{equation}
	[E(t_1,w_1) \geqslant E(t_2,w_2)] \iff \forall D \in \{D_i\}, [E(t_1,w_1)E(t_2,w_2)D = E(t_2,w_2)D]
\end{equation}
Since a SE erases a fraction of the terms in a document, every document defines a natural valuation for SEs on documents which is simply the count of unerased terms in a transformed document.  This will be represented with vertical bars $|\cdot|$ 
\begin{equation}
	V_D(E(t,w)) = |E(t,w)D|
\end{equation}
We can also define a formula analogous to (\ref{eq:ProbMeasureProjector}) defined by a set of weights and a set of documents $\{\omega_i,D_i\}$
\begin{equation}\label{eq:ProbMeasureEraser}
	V(E(t,w)) = \sum \omega_i \frac{|E(t,w) D_i|}{|D_i|}
\end{equation}

An intuition that will guide this work is that of the point-of-view-oriented user.  A user that is making a shallow reading of a text will expect only familiar terms and patterns, and will have a diminished ability to discern others that he or she does not expect.  We will assume here that a topical point of view will be associated to sets of lexical propositions that are both likely and rich in order relations.

\section{Conditionals for SEs}\label{sec:Valuations}
 \subsection{Material Implication}
 Using the concepts explained in the last section, we can start defining conditionals for SEs.  Material implication, for example, is defined as:
 \begin{equation}
	 (A \Rightarrow_m B) = (\lnot A) \lor B
 \end{equation}
 Two properties of probability measures can be used to evaluate a probability measure for this implication:
 \begin{equation}
	 \begin{matrix}
	 	V(\lnot A) = 1 - V(A) \\
		V(A \lor B) = V(A) + V(B) - V(A\land B)
	\end{matrix}
 \end{equation}
 Within a single document, the probability measure would then be:
 \begin{multline}
	 V_D(E(a,w_a) \Rightarrow_m E(b,w_b)) = 1 - V(E(a,w_a)) + V(E(a,w_a)\land E(b,w_b)) =\\
	 = 1 - \frac{|E(a,w_a)D|}{|D|} + \min_{[E(c,w_c)\geqslant_D E(a,w_a)]\land[E(c,w_c)\geqslant_D E(b,w_b)]}\frac{|E(c,w_c)D|}{|D|}
 \end{multline}
 This formula has all the known problems of material implication, like that of being 1 whenever $E(a,w_a)$ annihilates the document completely, so it will give probability 1 to documents without any occurrence of $a$ or $b$.   We have used a particular probability measure to avoid the cumbersome interpretation of what a meet and a join of SEs are.   Strictly speaking, a join $E_1 \lor E_2$ would be a transformation including both $E_1$ and $E_2$.  Within a single document a SE can always be found (even though it will very likely not be unique), but for a set of documents, the existence of join and meet defined in this way is not guaranteed.  

 \subsection{Subjunctive Conditional}
 A much more useful probability is that of the subjunctive (Stalnaker) conditional $\hspace{2pt}\Box\hspace{-4pt}\rightarrow$.   The base for computing this is the Ramsey test, which starts by assuming the antecedent as true with a minimum change of beliefs.   In this work we interpret that as taking the document transformed by the ``antecedent'' eraser $E(a,w_a)D$ as the whole document, and then compute the fraction of it that would be preserved by the further application of the ``consequent'' eraser $E(b,w_b)(E(a,w_a)D)$.  This produces a formula resembling a conditional probability:
 \begin{equation}
	 V_D(E(a,w_a) \hspace{2pt}\Box\hspace{-4pt}\rightarrow E(b,w_b)) = P_D(E(b,w_b)|E(a,w_a)) = \frac{|E(a,w_a)E(b,w_b)D|}{|E(a,w_a)D|} 
 \end{equation}
 This number will be 1 when $E(b,w_b) \geqslant E(a,w_a)$, and will be between 0 and 1 whenever $|E(a,w_a)D|\neq 0$.

 This formula still has problems when $a$ is not in the document, because in that case both $|E(a,w_a)E(b,w_b)D| = 0$ and $|E(a,w_a)D| = 0$.  A standard smoothing technique can be used in this cases using averages on a whole collection or estimates of them:
 
 \begin{equation}
	 \begin{matrix}
		 |E(a,w_a)E(b,w_b)\tilde{D}_0| = |E(a,w_a)E(b,w_b)D_0| + \mu|E(a,w_a)E(b,w_b)D_{avg}|\\
		 |E(a,w_a)E(b,w_b)\tilde{D}_0| =  |E(a,w_a)D_0| + \mu|E(a,w_a)D_{avg}|
	\end{matrix}
\end{equation}

Conditional probability when the terms are not present in an actual document would be $\frac{|E(a,w_a)E(b,w_b)D_{avg}|}{|E(a,w_a)D_{avg}|}$.  This value should be given the interpretation of ``undetermined''.

The final formula proposed for the probability of implication is then:

\begin{equation}\label{eq:UncertainConditional}
	 P_D(E(a,w_a) \hspace{2pt}\Box\hspace{-4pt}\rightarrow E(b,w_b)) = \frac{|E(a,w_a)E(b,w_b)D|+\mu|E(a,w_a)E(b,w_b)D_{avg}|}{|E(a,w_a)D|+ \mu|E(a,w_a)D_{avg}|}
 \end{equation}

\subsection{Topic-Specific Lattices}
If we think of a user going through a text document in a hurried and shallow way, we may assume that his attention will be caught by familiar terms, and then he or she will get an idea of the vocabulary involved that is biased towards the distribution of terms around these familiar set.

Suppose we take a set of SEs with a fixed width centred in different (but semantically related) terms.  We will assume that the pieces of text preserved by these can be thought as a lexical representation of the topic.  In this text, we can look for order relations between narrower SEs centred in the same terms or others, as a representation of the document.

If a text is very long, or there are a large number of documents taken as a corpus to characterise lexical relations in a topic, it is not convenient to require strict conditions like $E(a,w_a)E(b,w_b)D = E(b,w_b)D$ for al large document $D$ or for all documents $D_i$ in a large set, because then recognised order relations would be very scarce.  A more sensible approach would be to assess a probability within the text preserved by the SEs that define the topic, which would be:
\begin{multline}
	P_{topic}(E(a,w_a) \hspace{2pt}\Box\hspace{-4pt}\rightarrow E(b,w_b)) =\\
	= \max_{k_i}\left( P_{topic}([E(k_i,w_t)E(a,w_a)] \hspace{2pt}\Box\hspace{-4pt}\rightarrow [E(k_i,w_t)E(b,w_b)])\right)
\end{multline}

Restricting ourselves to the set of keywords $\{k_i\}$, the maximum value would always be for the topic-defining SE with the same central term as the antecedent SE $E(a,w_a)$ ($a = k_i$), which simplifies the formula to:
\begin{multline}\label{eq:TopicalUConditional}
	P_{topic}(E(a,w_a) \hspace{2pt}\Box\hspace{-4pt}\rightarrow E(b,w_b)) =\\
	= \frac{|E(a,w_a)E(b,w_b)E(a,w_t)D|+\mu|E(a,w_a)E(b,w_b)E(a,w_t)D_{avg}|}{|(E(a,w_a)D|+ \mu|E(a,w_a)D_{avg}|}
\end{multline}
for any $w_a < w_t$, where $w_t$ is the width of the SEs used to define the topic.  For large values of $w_t$ this would be equivalent to general formula (\ref{eq:UncertainConditional}).

\section{An Example}\label{sec:examples}
A particular topic might define its own particular \textit{sub-language}; this is a well known fact, and an interesting matter for research \cite{Harris1982discourse}.  The differences between these sub-languages and the complete language have been studied for very wide topics, such as scientific research domains \cite{BiomedicalSublanguages}.  In this work, we will aim to much more fine-grained topics, which could be found dominating different parts of a single document. Fine-grained sub-languages such as those would not depart from the whole language of the document significantly enough to be described grammatically or semantically as a sub-language in its own right, but will be rather a preference of some lexical relations over others.

As an illustration of how SE-based Uncertain Conditionals can be used to explore and describe the use of language characteristic of a particular, fine-grained topic, we will use two versions of a single document in different languages, and find the relations between terms chosen to define a topic.   We have chosen the literary classic novel Don Quixote as the subject for examining lexical features.  Two versions were used of this novel: the original in spanish \cite{QuixoteSpanish}, as it has been made available by project Gutenberg, and a translation to English by John Ormsby, obtained from the same site \cite{QuixoteEnglish}.
\begin{table}[h]
	\centering
	\begin{tabular}{|c|c|c|}\hline
		language & No. of tokens & No. of terms \\ \hline
		Spanish & 387675 & 24144 \\ \hline
		English & 433493 & 15714\\ \hline
	\end{tabular}\vspace{3pt}\\
	\caption{Characteristics of the Spanish and English version of don Quixote as a plain text sequence}
	\label{tab:QuixoteData}
\end{table}
In this text, we define a topic by the keywords $\{sword,hand,arm,helmet,shield\}$ and their Spanish equivalents $\{espada, mano, brazo, yelmo, adarga\}$ and the width for the topic-defining SEs was chosen to be 10.  Co-occurrence studies have found that the most meaningful distances between terms are from 2 to 5 \cite{HAL}, so we took twice the highest recommended co-occurrence distance to capture also relations between terms \textit{within} non-erased windows.  Information about the text and the topics is given in table \ref{tab:QuixoteData}.

Order relations were tested with formula (\ref{eq:TopicalUConditional}), and those implying the lower values of $w_a$ and $w_b$ (widths of antecedent and consequent) were taken as representative.  The values can be seen in table \ref{tab:TopicLattices}.
\begin{table}[h]
	\centering
	\begin{tabular}{|c||c|c|c|c|c|}\hline
		& sword & hand & arm & helmet & shield \\\hline \hline
		sword	& trivial	& P(2$\sqsupseteq$3)=87\%	&  P(1$\sqsupseteq$3)= 93\% & -	& P(8$\sqsupseteq$3) = 59\%\\\hline
		hand	& P(2$\sqsupseteq$ 3) = 96\%	& trivial	& P(2$\sqsupseteq$3)= 71\% & -	& - \\\hline
		arm	& P(2$\sqsupseteq$1)=96\% & P(2$\sqsupseteq$3)=87\%	& trivial & P(1$\sqsupseteq$3)=71\%	& P(3$\sqsupseteq$4) = 53\%\\\hline
		helmet	& -	& -	&  -	& trivial	& - \\\hline
		shield	&  P(7 $\sqsupseteq$3)=88\%	& - & P(3 $\sqsupseteq$3)=87\%	& - &  trivial \\\hline
	\end{tabular}
	\begin{tabular}{|c||c|c|c|c|c|}\hline
		& espada & mano & brazo & yelmo & adarga \\\hline \hline
		espada	& trivial	& P(4$\sqsupseteq$3)=67\%	& P(6$\sqsupseteq$3)= 85\% & -	& P(2$\sqsupseteq$7) = 52\%\\\hline
		mano	& P(2$\sqsupseteq$ 3) = 89\%	& trivial	& P(4$\sqsupseteq$3)= 75\% & -	& P(4$\sqsupseteq$3)= 63\% \\\hline
		brazo	& P(5$\sqsupseteq$3)=89\% & P(3$\sqsupseteq$3)=94\%	& trivial & -	& P(1$\sqsupseteq$3) = 74\%\\\hline
		yelmo	& -	& -	&  -	& trivial	& - \\\hline
		adarga	&  P(6 $\sqsupseteq$3)=94\%	& P(3 $\sqsupseteq$3)=94\% & - & - &  trivial \\\hline
	\end{tabular}\vspace{3pt}\\
	\caption{Order relations between SEs with the lower values of window width, within a topic defined by a set of erasers of width 10 centred in the same 5 words, both in their English and Spanish version. Relations ($N_1\sqsupseteq N_2$) represent relations $E(t_{row},w_1)\hspace{2pt}\Box\hspace{-5pt}\rightarrow E(t_{column},w_2)$}
	\label{tab:TopicLattices}
\end{table}
 \subsection{Anomalies in the Ordering}
\begin{figure}[h]
	  \begin{center}
		  \includegraphics[width=4cm]{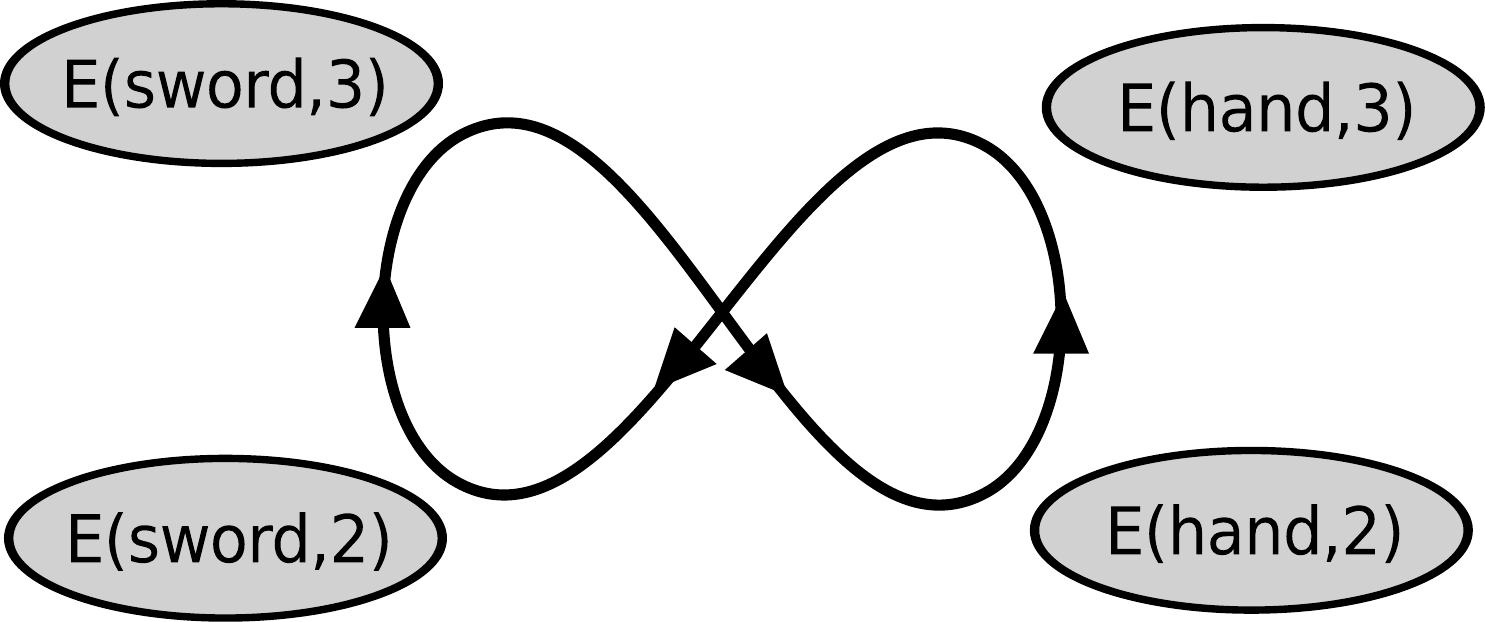}
	  \end{center}
	  \caption{Anomalous ordering of four SEs in the English topical lattice}
	  \label{fig:NonOrder}
  \end{figure}

 Table \ref{tab:TopicLattices} shows apparently paradoxical results.   Relations $E(sword,2)\hspace{2pt}\Box\hspace{-5pt}\rightarrow E(hand,3)$ and $E(hand,2) \hspace{2pt}\Box\hspace{-5pt}\rightarrow E(sword,3)$, both with probabilities above $87\%$, do not fulfill the properties of an order relation when considered together with $E(sword,3)\hspace{2pt}\Box\hspace{-5pt}\rightarrow E(sword,2)$ and $E(hand,3)\hspace{2pt}\Box\hspace{-5pt}\rightarrow E(hand,2)$ (see figure \ref{fig:NonOrder}).  This is a result of putting together partially incompatible scenarios: $E(sword,2)\hspace{2pt}\Box\hspace{-5pt}\rightarrow E(hand,3)$ is evaluated in the text preserved by $E(sword,10)$ and $E(hand,2) \hspace{2pt}\Box\hspace{-5pt}\rightarrow E(sword,3)$ is evaluated in the text preserved by $E(hand,10)$.
 
  Anomalies in the order can be resolved by simply choosing some of the relations on the basis of their higher probability (in this case, $E(hand,2) \sqsupseteq E(sword,3)$ with 96\% over $E(sword,2) \hspace{2pt}\Box\hspace{-5pt}\rightarrow E(hand,3)$ wiwth 87\%, or collapsing the involved SEs in a class of equivalence, so the inconsistency is removed.

 \subsection{Lattices for Two Languages}
 The sets of relations obtained are strikingly similar for the two languages, with more differences polysemic terms like ``arm'' (which appears in spanish with different terms for its noun meaning and for its verb meaning) and ``sword'' which corresponds to different kinds of weapons with their own name in Spanish, from which ``sword'' is just the most frequent.  Moreover, the anomaly in the orderings of SEs centred in ``sword'' and ``hand'' does not appear between their spanish counterparts ``espada'' and ``mano'', but is replaced by a very similar pair of relations.

 This kind of analysis provides a promising way of finding regularities between different languages, or even analogies between different terms in the same language.  It is easy to isolate the transformations needed to go from the English Lattice to the Spanish one, as a lattice morphism.   The differences of both could even suggest a valuation, a mapping to a simpler common lattice.

\section{Discussion and Conclusion}\label{sec_conclusion}

In this work, we have shown how the framework of SEs provides a natural platform to define logical relations resembling those employed in Boolean logics, but also more complex ones, like the subjunctive conditional.    Quantitative implementation follows naturally from the parallel between lexical measurements and quantum ideal measurements, producing a formula that is both simple and easy to compute for concrete cases.

The proposed formula also allows to include relations restricted to only a chosen bit of the text, that surrounding the occurrences of keywords.   This allows to extract relations between terms that can be expected to be characteristic of the text about a particular topic.

The proposed formula was applied to a simple example, with very interesting results.   Two main features can be observed in the results:
\begin{enumerate}
	\item Anomalies can appear in the resulting order relation, coming from the existence of transformations that are incompatible in the sense of quantum incompatibility.   These can be removed easily if a proper lattice-valued representation is to be obtained, but can also be studied as an evidence of useful patterns as well.
	\item The relation structures between SEs make visible common features of the representation of a text in different languages: terms that mean something similar will be embedded into similar patterns of relations.
\end{enumerate}

As a matter for future research, both observations can be explored further: the causes and characteristics of the anomalies in order relations between SEs as assessed by uncertain conditionals, and the possibility of putting the multi-language representation in terms of morphisms between lattices of SEs.

In particular, having similar lattices for two versions of the same text in different languages invites to find an optimal way of defining a common valuation that would assign both lattices to a simpler third lattices with their common features.   This, in particular, is a very promising direction of research, and a novel approach to multi-lingual text processing.

\section*{Acknowledgements}
 This work was developed under the funding of the Renaissance Project ``Towards Context-Sensitive Information Retrieval Based on Quantum Theory: With Applications to Cross-Media Search and Structured Document Access'' EPSRC EP/F014384/1, School of Computing Science of the University of Glasgow and Yahoo! (funds managed by Prof. Mounia Lalmas).
\bibliographystyle{abbrv}

\end{document}